\documentclass[sigconf,nonacm]{acmart}

\DeclareMathAlphabet\mathbfcal{OMS}{cmsy}{b}{n}
\usepackage[utf8]{inputenc}

\usepackage{graphicx}
\graphicspath{ {./images/} }
\usepackage{color}
\usepackage[x11names, table]{xcolor}
\usepackage{algorithm,algcompatible}
\algnewcommand\algorithmicreturn{\textbf{return}}
\algnewcommand\RETURN{\algorithmicreturn}
\algnewcommand\algorithmicprocedure{\textbf{procedure}}
\algnewcommand\PROCEDURE{\item[\algorithmicprocedure]}%
\algnewcommand\algorithmicendprocedure{\textbf{end procedure}}
\algnewcommand\ENDPROCEDURE{\item[\algorithmicendprocedure]}%
\algnewcommand{\algvar}[1]{{\text{\ttfamily\detokenize{#1}}}}
\algnewcommand{\algarg}[1]{{\text{\ttfamily\itshape\detokenize{#1}}}}
\algnewcommand{\algproc}[1]{{\text{\ttfamily\detokenize{#1}}}}
\algnewcommand{\algassign}{\leftarrow}
\usepackage{algpseudocode}
\usepackage{graphicx}
\usepackage{booktabs}
\usepackage{longtable}
\usepackage{caption}
\usepackage{subcaption}
\usepackage[normalem]{ulem}
\usepackage{ulem}
\useunder{\uline}{\ul}{}
\usepackage{tabularx}

\usepackage{amssymb}
\usepackage{amsmath}
\newcommand{\fastman}{{\fontfamily{lmtt}\selectfont FaSTM$\forall$N}}

\newcommand{\exstraqt}{{\fontfamily{lmtt}\selectfont ExSTraQt}}

\usepackage{pifont}

\usepackage{rotating}
\usepackage{multirow}
\PassOptionsToPackage{hyphens}{url}\usepackage{hyperref}
\usepackage{hyperxmp}
\usepackage{tikz}
 
\usepackage{wrapfig}
\usepackage{float}
\usepackage{orcidlink}

\begin{document}
\title{Extracting Money Laundering Transactions from Quasi-Temporal Graph Representation}
\author{Haseeb Tariq~\orcidlink{0000-0003-0756-3714}}
\affiliation{
  \institution{ING Bank}
  \city{Amsterdam}
\country{The Netherlands}
}
\affiliation{
  \institution{Eindhoven University of Technology}
  \city{Eindhoven}
\country{The Netherlands}
}
\email{m.h.tariq@tue.nl}
\author{Marwan Hassani~\orcidlink{0000-0002-4027-4351}}
\affiliation{%
  \institution{Eindhoven University of Technology}
    \city{Eindhoven} 
\country{The Netherlands}
}
\email{m.hassani@tue.nl}

\renewcommand{\shortauthors}{H. Tariq and M. Hassani}
\renewcommand{\authors}{Haseeb Tariq and Marwan Hassani}

\begin{abstract}

Money laundering presents a persistent challenge for financial institutions worldwide, while criminal organizations constantly evolve their tactics to bypass detection systems. Traditional anti-money laundering approaches \textit{mainly} rely on predefined risk-based rules, leading to resource-intensive investigations and high numbers of false positive alerts. In order to restrict operational costs from exploding, while billions of transactions are being processed every day, financial institutions are investing in more sophisticated mechanisms to improve existing systems. In this paper, we present {\exstraqt} (\underline{Ex}tract \underline{S}uspicious \underline{Tra}nsactions from \underline{Q}uasi-\underline{t}emporal Graph Representation), an advanced supervised learning approach to detect money laundering (or suspicious) transactions in financial datasets. Our proposed framework excels in performance, when compared to the state-of-the-art AML (Anti Money Laundering) detection models. The key strengths of our framework are sheer simplicity, in terms of design and number of parameters; and scalability, in terms of the computing and memory requirements. We evaluated our framework on transaction-level detection accuracy using a \textit{real} dataset; and a set of synthetic financial transaction datasets. We \textit{consistently} achieve an uplift in the F1 score for most datasets, up to \textbf{1\%} for the real dataset; and more than \textbf{8\%} for one of the synthetic datasets. We also claim that our framework could \textit{seamlessly} complement existing AML detection systems in banks. Our code and datasets are
available at https://github.com/mhaseebtariq/exstraqt.

\end{abstract}

\ccsdesc[500]{Computing methodologies~Batch learning}
\ccsdesc[500]{Theory of computation~Network flows}
\ccsdesc[300]{Computing methodologies~Massively parallel algorithms}

\keywords{Money Laundering Detection, Graph Modeling, Transaction Monitoring, Distributed Computing}

\maketitle

\section{Introduction}
\label{sec:introduction}

The rise of digitization and the advent of new technologies are shaping a new financial infrastructure, which allows faster; interconnected; and increasingly real-time payments \cite{mckinseyreport}. An average bank processes billions of transactions every year, offering multiple payment methods based on customer needs. Money launderers exploit the same financial ecosystem to integrate their illegally obtained proceeds. The consequences are immense. Annually, between 2 and 5\% of the global GDP is laundered \cite{unodc}. More importantly, money laundering promotes crime and corruption; and has negative consequences on growth rates, money demand, income distribution, and tax revenues.

Money laundering techniques typically employ a complex series of payments and cash flow schemes across several banks, operating in distant jurisdictions \cite{mlfiu}. The process can be divided into three main stages. The first stage consists of the introduction of illegally acquired funds into the financial system, which is often referred to as \textit{placement}. Once this is achieved, money launderers use \textit{layering} to obfuscate the trace of the funds by moving them away from the source. The third and last step is the \textit{integration} of those funds in the mainstream economy, which is mostly achieved by (again) abusing financial services.

Central banks and the overarching regulatory authorities around the world compel financial institutions to maintain integrity in the services they provide. Maintaining that integrity involves preventing the abuse of the financial system for activities such as money laundering and terrorist financing. To comply with these requirements, financial institutions implement processes and procedures to mitigate the risks of abusive practices. The alerts generated by the transaction monitoring systems are generally risk-based. Those alerts are then manually investigated by analysts or trained AML (Anti Money Laundering) experts. A monitoring system generally includes a set of predefined business rules, consisting of scenarios with threshold values. These rules drive the characterizations of the customers and their transactions. In addition to rule-based systems, financial institutions are increasingly using more advanced mechanisms \cite{industryperspective}. This is essential because banks have to deal with ever increasing volumes of data, while the nature of financial fraud is becoming more intricate over time.

Despite the deployments of these systems, according to some statistics \cite{amlstats}, on average 95\% of \textit{automatically} generated alerts are false positives, while most suspicious activities go undetected. Banks incur huge costs by reviewing thousands of alerts every week, with the \textit{main} objective of eliminating a huge number of false alerts. Compliance operations costs must also be taken into account since the human (in-the-middle) component is necessary to \textit{fairly} conclude complex investigations.

To address these challenges, in this paper, we propose a supervised machine learning framework to effectively identify anomalous transactions in financial datasets. Our key contributions are as follows:
\begin{itemize}
    \item A rich set of graph-based transaction features, tailored towards AML and suspicious transactions detection, for a supervised learning model
    \item Massively parallelizable implementation of complex subgraph metrics
    \item A novel method for quantifying \textit{flow}-based money laundering activities
\end{itemize}

Section \ref{sec:related} discusses existing work. Sections \ref{sec:preliminaries} \& \ref{sec:method} introduce the definitions and {\exstraqt}, respectively. Section \ref{sec:experiments} details the experimental evaluation; and Section \ref{sec:conclusion} concludes the paper.
\section{Related Work}
\label{sec:related}
When it comes to TM (transaction monitoring) and AML modeling, graph is a natural choice for data representation. By connecting the source of a fund transfer with its target, the transactional behavior of the participating entities can be captured on a broader scope. The history surrounding a single transaction; the direction; and the flow of money also carry important contextual information. We are now going to review the most relevant developments in the money laundering detection techniques, with a focus on graph-centric algorithms.

Recent works have leveraged graph-based features to improve the performance of \textit{traditional} detection methods. For example, the (GFP) Graph Feature Preprocessor \cite{gfplib} is a library used to produce a rich set of transaction features, for a downstream supervised learning task. Gradient boosting-based models, such as \cite{xgb, lgbm}, are particularly effective when used in combination with these features. We use GFP as one of our main benchmark methods.

Unsupervised approaches based on graphs target the broader transactional behaviors of customers. ReDiRect \cite{redirect} uses graph modeling to incrementally remove normal nodes from the search space, consequently, unearthing the money laundering nodes. Flowscope \cite{flowscope} searches for dense flows (or transfers) of funds in multipartite transaction graphs, while employing a custom metric to effectively identify (and quantify) flows. \cite{starnini} proposes a practical data pipeline to detect smurf-like subgraphs. Their framework is based on joins of sequential transactions, constrained by a time-window and filtering thresholds. The main limitation of these methods is that they only target specific motifs. Although they are quite practical in the sense that they can complement existing techniques.

Other methods focus on community detection, assuming that money launderers will show \textit{interdependent} flows or strong relationships with each other. One of these frameworks \cite{intsolution} identifies money laundering groups within temporal networks. Maximally connected subgraphs are extracted from graphs of \textit{connected} (fund) transfers. Leveraging domain knowledge, some edges are discarded, retaining only relevant components. The filtered components are then decomposed through community detection; and each community is ranked with a score that incorporates risk factors. {\fastman} \cite{fastman} is an innovative approach that builds on the aforementioned methods but provides a unique contribution. Transactions are mapped as nodes in a directed graph, representing a higher-order abstraction. The transaction chains are constructed on the basis of a time window, connecting each transaction to subsequent (and relevant) transactions within the period. The temporal graph is then partitioned by employing a community detection algorithm; and the resulting communities are ranked and analyzed through (undisclosed) AML risk criteria. These approaches highlight the efficacy of unsupervised learning techniques in the domain. However, these methods are based on \textit{deterministic} steps, while the inherent money laundering patterns might be far more complex. Based on industry perspectives \cite{industryperspective}, the shortcomings of existing transaction monitoring systems are catalyzing the adoption of more sophisticated machine learning technologies.

With the advancements in GNN (Graph Neural Network) and GT (Graph Transformers) models, these architectures are also being intensively researched in the context of AML modeling. One of their primary applications is to encode the graph structure in a lower dimension to improve downstream (training) tasks. In \cite{amlbitcoin} the authors experiment with Graph Convolutional Networks (GCNs), to enhance the representation of a transaction graph. They concatenate embeddings, obtained by GCN, with original features to increase the accuracy of a supervised learning model. In another study, \cite{ibmsynth} demonstrates that GNN models can recognize laundering patterns in extremely imbalanced datasets. \cite{ppgnn} further adapts GNNs for directed multigraphs by introducing techniques such as multigraph port numbering, ego IDs, and reverse message passing, allowing the model to capture more complex money laundering patterns. LaundroGraph \cite{laundrograph} is a fully self-supervised method for AML detection. Financial interactions are represented as an attributed customer-transaction bipartite graph. The architecture consists of a GNN encoder trained on a user profile feature set; and a feed-forward decoder built for the task of anomaly detection. FraudGT \cite{fraudgt} is the most recent graph-based transformer method. In their paper, the authors compare their model performance with a variety of GNN-based models \cite{gnn-1, gnn-2, gnn-3, ppgnn, gnn-5, gnn-6, gnn-7, gnn-8, gnn-9, gnn-10}. They prove the superiority of their model over all these methods. FraudGT \cite{fraudgt} and MultiGNN \cite{ppgnn} are also two of our main competing benchmarks.

We believe that while these advanced neural network-based models are quite promising; the same, or even better, levels of performance can be achieved by creatively engineering features for much simpler machine learning algorithms. The benefits of such an approach are three-fold:
\begin{enumerate}
    \item A tree-based boosting model, such as XGBoost \cite{xgb}, provides better interpretability, compared to deep neural networks. In a highly regulated industry such as financial services, this is extremely desirable.
    \item The design complexity of neural networks also does not help with the wide-spread adoption in the industry.
    \item Most GNN architectures are still not scalable for practical real-world applications. Transformer-based models such as FraudGT are quite scalable, although they come with a very high cost of compute and memory resources.
\end{enumerate}
\section{Preliminaries}
\label{sec:preliminaries}

In this section, we will briefly introduce some of the basic preliminary concepts and definitions required to understand the AML (Anti Money Laundering) problem space.

\subsubsection{Financial transaction data}
{\exstraqt} is designed for financial datasets which include: the account identifiers of the main party and the counter-party; the amount being exchanged between those parties; and the timestamp of the activity. The timestamp information is specifically important to capture temporal relationships in transactional interactions. The framework can be \textit{easily} adapted with the availability of richer transaction metadata.

\subsubsection{Real and \textit{injected} anomalies}
Money Laundering does not always adhere to specific or recognizable patterns. Recently, the IBM Watson research lab has attempted to model \cite{ibmsynth} the three stages of money laundering - placement, layering, and integration - as a set of known patterns or suspicious typologies. We will use these labeled synthetic datasets, where transactions will be predicted as suspicious or legitimate. For the real dataset, we will use the openly available Ethereum Phishing Data \cite{ethereum1, ethereum2}. Here, the labels are the phishing attempts made by either the source; or the destination account.

\subsection{Definitions}

\subsubsection{Transaction dataset}  
A transaction dataset \(D \in \mathbb{R}^{n \times k}\) is a matrix where each of the \(n\) rows corresponds to a financial transaction; and each of the \(k\) columns corresponds to its attributes.

\subsubsection{Transaction graph}  
The dataset \(D\) can be represented as a static directed (transaction) graph \( G = (V, E, X) \), where \(V = \{v_1, \ldots, v_m\}\) represents financial accounts as nodes; \(E\), the set of directed edges \(e_{s,t} = (v_s, v_t)\), and \(X\), the edge attribute matrix containing vectors \(x_{s,t}\) with transaction details. The subscript $s$ represents the source account that sends the amount; and $t$, the target account that receives the amount.

We also use an \textit{aggregated} version of the \textit{directed graph}. For that version, an aggregated (and unique) edge between $s, t \in \mathcal{V}$ is represented as $(s \rightarrow t) \in \mathcal{E}$. The total amount transferred from $s$ to $t$ is represented as an edge attribute, $\mathcal{A}_{s,t}$. The two node attributes, $\mathcal{S}$ and $\mathcal{R}$, represent the total amount sent and the total amount received by a node, respectively. The weight $\mathcal{W}$ for an aggregated edge is calculated as follows:
\begin{equation}
\label{eq:weight}
   \mathcal{W}(s \rightarrow t) = \frac{\mathcal{A}_{s \rightarrow t}}{\mathcal{S}_s} + \frac{\mathcal{A}_{s \rightarrow t}}{\mathcal{R}_t} 
\end{equation}

This weight serves as the main driver for constructing the communities described in Section \ref{ref:comm-det}. The addition of the two ratios, one from the point of view of the sender and the other from the point of view of the receiver, ensures that a money launderer can not easily manipulate the system. For example, agents in a network can avoid appearing together (in a community) by using a high-volume intermediary account. In such a situation, only one of the two terms in Equation \ref{eq:weight} can be minimized, not both.

\subsubsection{Money laundering account types}
\label{ref:ml-acc-types}
\begin{figure}
    \centering
    \includegraphics[width=0.7\linewidth]{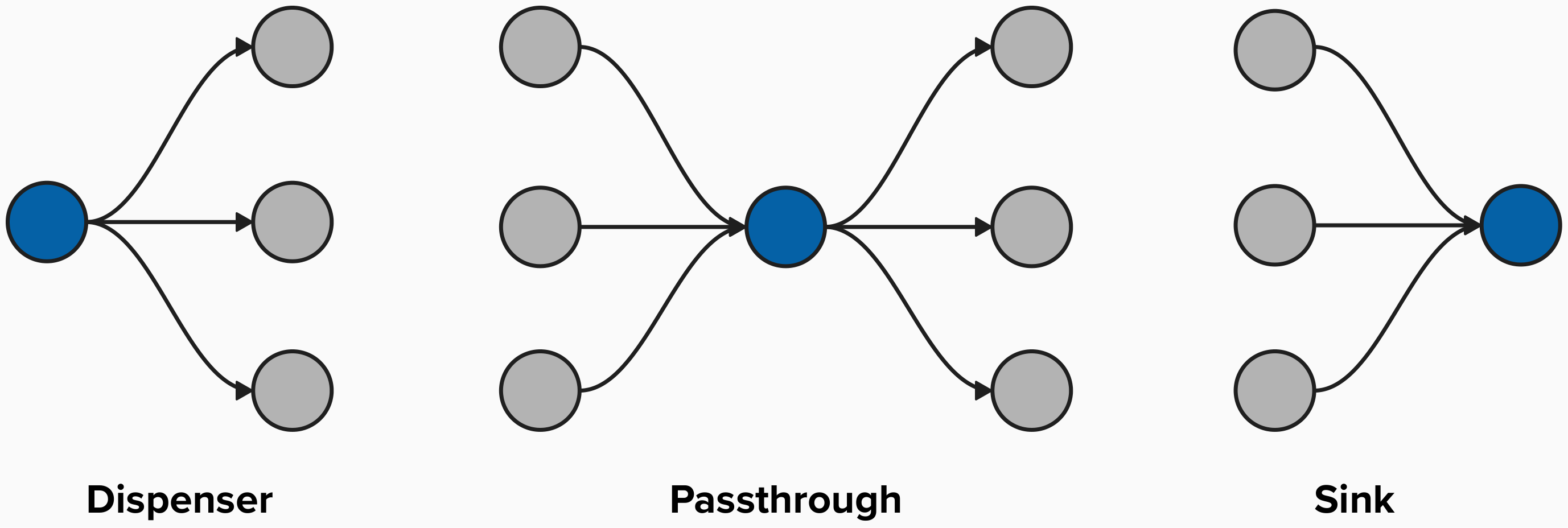}
    \caption{Types of ML flow accounts. {\color[HTML]{1434A4} \textbf{Blue}} node represents the node that is being defined.}
    \label{fig:flow-types}
\end{figure}

For the (money) \textit{flow}-based features that we define in Section \ref{ref:flow-quant}, we want to introduce the concept of money-laundering account types, as shown in Figure \ref{fig:flow-types}.
\begin{itemize}
    \item \textit{Dispenser}: Represents the source accounts from which the flow of money laundering funds are initiated. These types of accounts are generally used (for example, by drug mules) in the \textit{placement} phase.
    \item \textit{Passthrough}: Represents the accounts used (for example, by shell companies) to obfuscate the money trail. These are generally used in the \textit{layering} phase.
    \item \textit{Sink}: Represents the accounts at which the flow of money (laundering funds) terminates. These types of accounts are generally used in the \textit{integration} phase.
\end{itemize}

\subsection{Challenges}
Transaction labels are highly imbalanced. Usually, legitimate transactions outnumber illicit ones, which appear in ratios as low as 0.05\% (see Table \ref{tab:dataset-stats}). Models might therefore be biased towards the detection of the majority class. To further complicate things, there is often a degree of overlap between the normal class and the laundering class. \textit{Concept drift} is another challenge, where the statistical properties of the target variable change over time. Finally, poor data quality and complicated data models are significant challenges that financial institutions also struggle to solve \textit{completely}.
\section{Our Method: {\exstraqt}}
\label{sec:method}
\begin{figure*}[tp]
    \centering
    \includegraphics[width=0.75\linewidth]{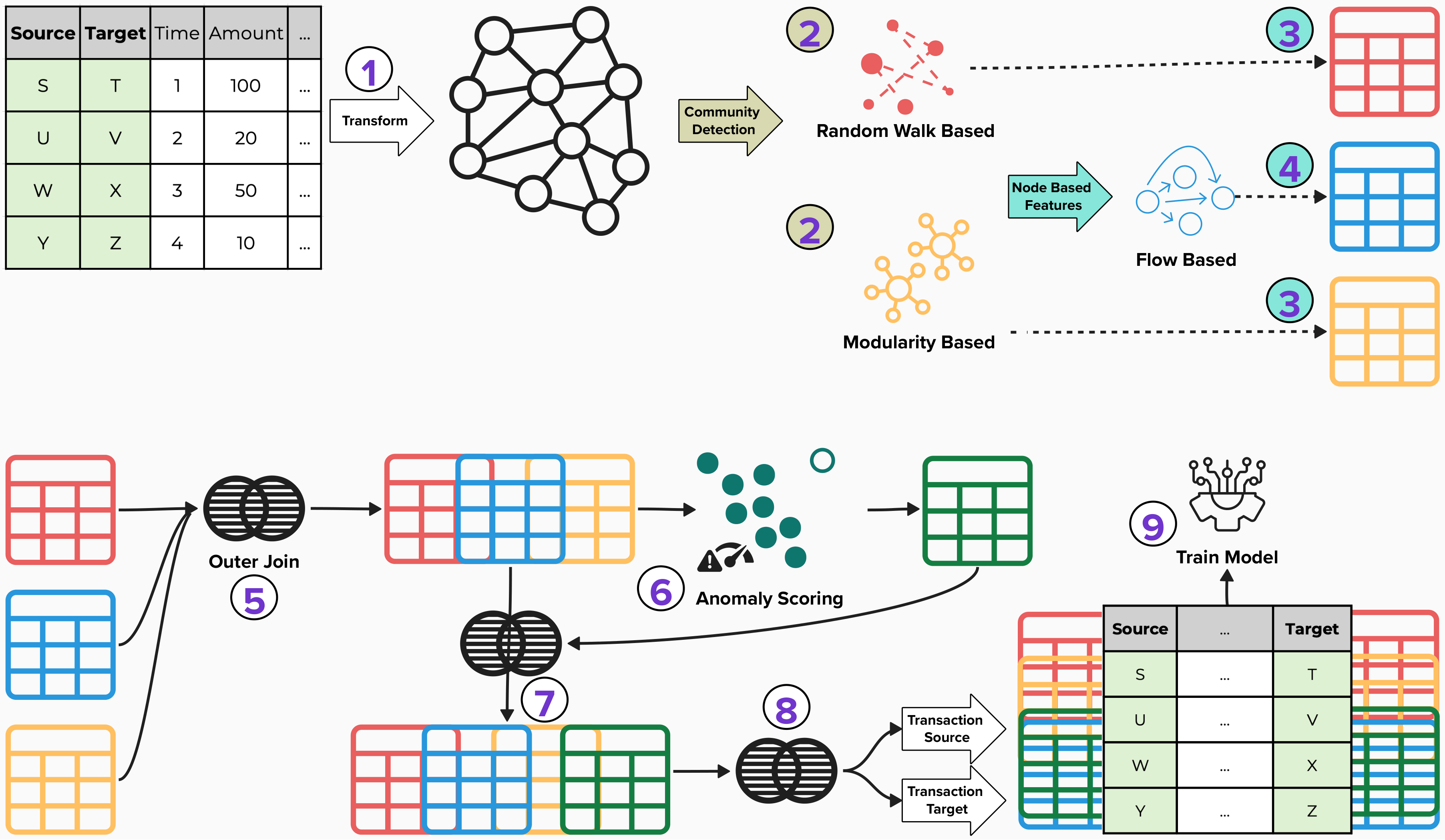}
    \caption{Overview of the framework.}
    \label{fig:framework}
\end{figure*}

Our framework design consists of 9 \textit{sequential} stages, as shown in Figure \ref{fig:framework}. We want to stress here that, while having 9 stages, the design is extremely simple. There are no convoluted or strongly coupled dependencies among the different stages. In the following sections, we will go through each of the (crucial to explain) stages in detail.

\subsection{Community Detection}
\label{ref:comm-det}

To capture the behavioral patterns of the nodes (or accounts) at different levels, we employ the following two methods:
\begin{itemize}
    \item \textit{Top-down approach}: We use a modularity-based community detection method, the Leiden algorithm \cite{leieden}, to assign a \textit{global} context to the nodes under which they operate. The resulting communities are non-overlapping, i.e. one node can only appear in a single community.
    \item \textit{Bottom-up approach}: Here, we start from a single node and build a community (around it) using the random-walks concept from PageRank \cite{pagerank}. The idea here is to capture the \textit{local} context of each node under which it operates. In the real world, a money laundering agent can facilitate multiple money laundering networks; therefore, an approach that results in overlapping communities is desirable here. We construct the communities for up to n-hops (in both directions). For the experiments, we used n = 2, which results in a maximum of 4-hop \textit{directed} community. The \textit{distributed} implementation of this algorithm is provided with the submitted code.
\end{itemize}

\subsection{Flow-based Features}
\label{ref:flow-quant}
\begin{figure}
    \centering
    \includegraphics[width=1\linewidth]{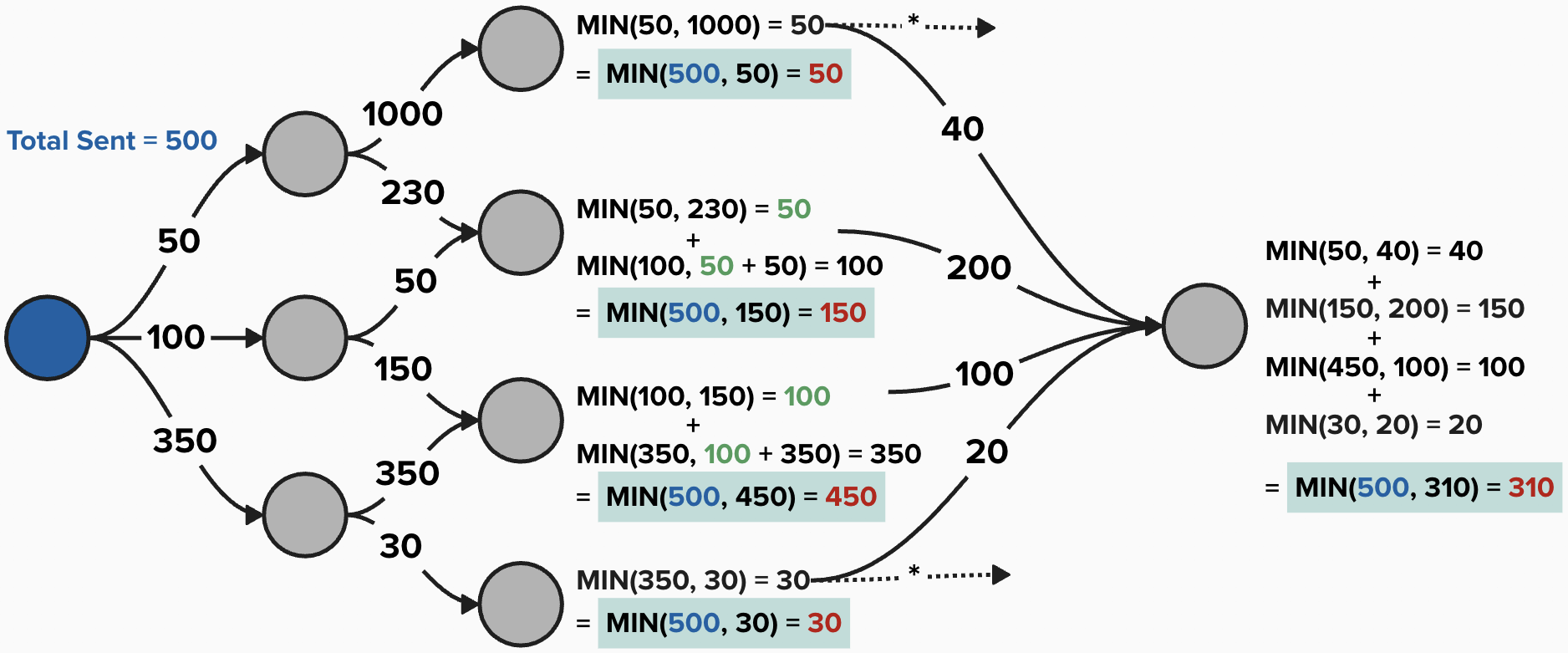}
    \caption{Example of \textit{dispense} flow calculation.}
    \label{fig:flow-example}
\end{figure}

As introduced in Section \ref{ref:ml-acc-types}; Algorithm \ref{alg:flow-normal} explains the detailed implementation of these features. We essentially want to quantify, based on the specific money laundering profile (see Figure \ref{fig:flow-types}): once a node sends or receives an amount, how much of it is carried forward; and to how many accounts. These features are estimated from the aggregated graph. Figure \ref{fig:flow-example} visually explains the same algorithm.

\begin{algorithm}
\begin{algorithmic}
\caption{Flow-based features implementation for dispense accounts}
    \State $\mathrm{number\_of\_hops} \gets 5$ **[adjust according to the risk appetite]
    \State $\mathrm{keep\_top\_n} \gets 50$ **[adjust based on domain knowledge]
    \State $\mathrm{total\_debit} \gets \mathrm{SUM} \mathopen{}\left( \mathrm{debits\_for\_node} \mathclose{}\right)$
    \State $\mathrm{prev\_edges} \gets \mathrm{node}.\mathrm{out\_neighbors}$
    \State $\mathrm{prev\_edges} \gets \mathrm{SORT\_DESC\_ON\_AMOUNT} \mathopen{}\left( \mathrm{prev\_edges} \mathclose{}\right)$
    \State $\mathrm{prev\_edges} \gets \mathrm{SLICE} \mathopen{}\left( \mathrm{prev\_edges}, \mathrm{keep\_top\_n} \mathclose{}\right)$
    \State $\mathrm{stats} \gets \mathopen{}\left[ \mathrm{GET\_AMOUNT\_STATS} \mathopen{}\left( \mathrm{prev\_edges}.\mathrm{target} \mathclose{}\right) \mathclose{}\right]$
    \For{$\mathrm{hop} \in \mathrm{range} \mathopen{}\left( 2, \mathrm{number\_of\_hops} + 1 \mathclose{}\right)$}
        \State $\mathrm{prev\_amounts} \gets \mathrm{MIN} \mathopen{}\left( \mathopen{}\left[ \mathrm{prev\_edges}.\mathrm{amount}, \mathrm{total\_debit} \mathclose{}\right] \mathclose{}\right)$
        \State $\mathrm{next\_edges} \gets \mathrm{SELF\_JOIN} \mathopen{}\left( \mathrm{prev\_edges}, \mathrm{left}.\mathrm{target} = \mathrm{right}.\mathrm{source} \mathclose{}\right)$
        \State $\mathrm{next\_edges}.\mathrm{amount} \gets \mathrm{MIN} \mathopen{}\left( \mathopen{}\left[ \mathrm{next\_edges}.\mathrm{amount}, \mathrm{prev\_amounts} \mathclose{}\right] \mathclose{}\right)$
        \State $\mathrm{next\_edges} \gets \mathrm{SORT\_DESC\_ON\_AMOUNT} \mathopen{}\left( \mathrm{next\_edges} \mathclose{}\right)$
        \State $\mathrm{prev\_edges} \gets \mathrm{SLICE} \mathopen{}\left( \mathrm{next\_edges}, \mathrm{keep\_top\_n} \mathclose{}\right)$
        \State $\mathrm{stats}.\mathrm{append} \mathopen{}\left( \mathrm{GET\_AMOUNT\_STATS} \mathopen{}\left( \mathrm{prev\_edges}.\mathrm{target} \mathclose{}\right) \mathclose{}\right)$
    \EndFor
\label{alg:flow-normal}
\end{algorithmic}
\end{algorithm}

\subsection{Flow-based Temporal Features}
\label{ref:flow-temporal}
Flow-based features can also be calculated taking into account the chronological order of the transactions. As we are now incorporating the temporal information into the logic, we will have to use the non-aggregated multigraph. To make Algorithm \ref{alg:flow-normal} work for this version, we have to make some adjustments; specifically, how we select $next\_edges$. In addition to the existing filter in $SELF\_JOIN$, we now need to add $left.timestamp <= right.timestamp$. In this way, we ensure that the transactions are only joined when they take place in chronological order.

\begin{figure}
    \centering
    \includegraphics[width=1\linewidth]{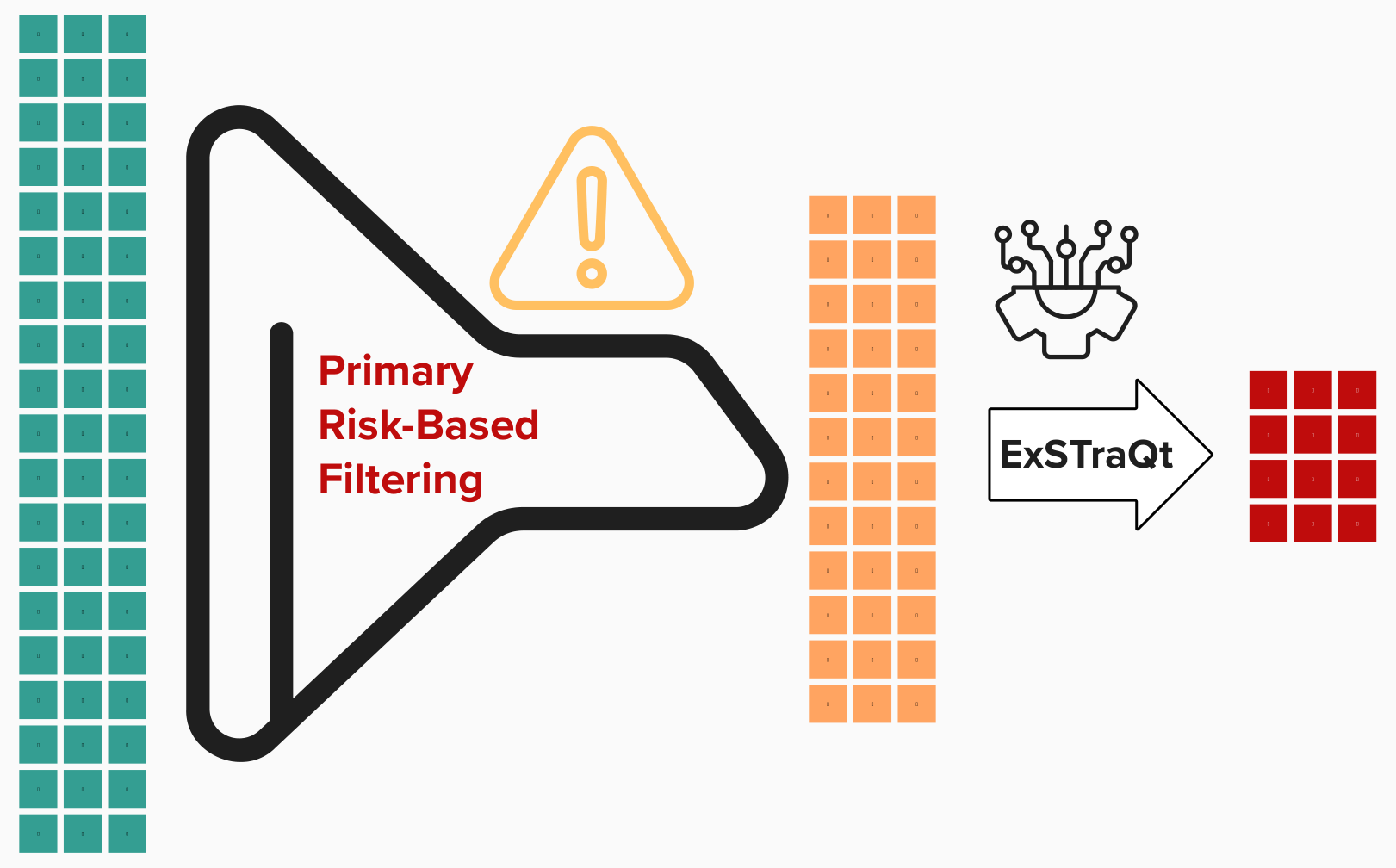}
    \caption{{\exstraqt} in production.}
    \label{fig:production}
\end{figure}

\subsection{Communities (or subgraphs) Features}
\label{ref:graph-features}
For each of the communities; and community types described in Section \ref{ref:comm-det}, we are now going to generate a rich set of graph-based features with the goal of capturing AML risks as comprehensively as possible.
\begin{itemize}
    \item \textit{Membership statistics}, such as, number of dispense; passthrough; and sink accounts
    \item \textit{Network properties}, such as degree distributions (in, out, and both), capturing fan-in and fan-out-like patterns; graph diameter, capturing distant relationships; degree assortativity; number of bi-connected components, capturing bi-partite relationships; and number of articulation points
    \item \textit{Turnover volumes}, such as, statistics for total debit/credit amounts per edge; and the turnover volume, based on the absolute difference of the total incoming and outgoing amounts in the network
    \item \textit{Temporal features}, as described in the next section
\end{itemize}

\subsubsection{Temporal features}
We transform the timestamp column in the transaction data into a trend feature by subtracting the earliest timestamp from each timestamp. Let us call this transformed time attribute, vector $\chi$. Then for every subgraph (or community), $\chi^c$ could represent the (time) feature vector for the transactions in community $c$. Finally, if vector ${\mathcal{A}^c}$ represents all the amounts per transaction in community $c$, we can define the weighted time feature for community $c$ as,

\begin{equation}
\label{eq:weighted-ts}
   TF(c) = \frac{\sum_{i=1}^{n} \mathcal{A}^c_i \chi^c_i}{\sum_{i=1}^{n} \mathcal{A}^c_i}
\end{equation}

In addition, we calculate statistics such as the weighted standard deviation and the weighted median using a similar approach. The idea here is to capture when; how frequently; or how sporadically do the community members interact with each other, while giving more weightage to the timestamps involving higher volumes.

\subsubsection{Distributed subgraph features generation}
The massively parallelizable implementation of the subgraph features generation is provided with the submitted code; and is explained in Algorithm \ref{alg:dist-feats}.

\begin{algorithm}
\begin{algorithmic}
\caption{Distributed subgraph features generation}
        \State $\mathrm{distributed\_file\_system} \gets \textrm{"<location>"}$
        \For{$\mathopen{}\left( \mathrm{node}, \mathrm{members} \mathclose{}\right) \in \mathrm{communities\_data}$}
            \State $\mathrm{sub\_graph} \gets \mathrm{graph}.\mathrm{INDUCED\_SUBGRAPH} \mathopen{}\left( \mathrm{members} \mathclose{}\right)$
            \State $\mathrm{community\_edges} \gets \mathrm{sub\_graph}.\mathrm{GET\_EDGES} \mathopen{}\left( \mathclose{}\right)$
            \State $\mathrm{community\_edges}.\mathrm{ADD\_FILE\_TO} \mathopen{}\left( \mathrm{distributed\_file\_system} \mathclose{}\right)$
        \EndFor
        \State $\mathrm{data} \gets \mathrm{READ} \mathopen{}\left( \mathrm{distributed\_file\_system} \mathclose{}\right)$
        \State $\mathrm{request} \gets \mathrm{PARALLEL\_MAP} \mathopen{}\left( \mathrm{data}, \mathrm{GET\_GRAPH\_FEATURES} \mathclose{}\right)$
        \State $\mathrm{results} \gets \mathrm{request}.\mathrm{REDUCE} \mathopen{}\left( \mathclose{}\right)$
\label{alg:dist-feats}
\end{algorithmic}
\end{algorithm}

\subsection{Anomaly Scoring}
\label{sec:anomaly}
Based on the node-level features that we have generated so far, we are going to train a simple anomaly detection model using IsolationForest \cite{isolationfor}. We are going to use the same features downstream for the supervised learning task, by joining them to the source and target accounts in the transaction data. The added advantage of acquiring an anomaly score per node is that (then) later, we can build \textit{interaction} features based on the source anomaly score and the target anomaly score for each transaction. In the ablation study, Section \ref{ref:ablation}, we will show how this step positively contributes to the overall accuracy of our model.

\subsection{Model Training}
After combining all the node-level features (including anomaly scores) from the previous steps into a single table, we join the resulting table with our main transaction table; once with the source account and then with the target account. The transaction table now has twice as many (additional) features as the nodes features. The final step is to simply feed the feature set to a supervised machine learning model. 

\subsection{{\exstraqt} in Production}
\label{ref:prod}
For banks, real-time detection of AML signals is kind of a non-requirement. In fact, without a man-in-the-middle alert validation process; and by relying too heavily on aggressive automation, banks can increase the chances of being fined by the regulators. The SAR (Suspicious Activity Report) filing requirements for most regional authorities are actually quite relaxed when it comes to meeting time deadlines. For example, \cite{sar} mentions a time window of 60 calendar days.

Keeping these requirements in mind, we propose \exstraqt, as a \textit{secondary} AML detection system trained in periodic batches; as opposed to the real-time system proposed in \cite{gfplib}. Figure \ref{fig:production} shows our proposed setup for {\exstraqt}'s deployment in a production environment. Essentially, our framework acts as a secondary detection system that takes as input the signals produced by a much more conservative \textit{primary} detection system, for example, a system built on AML risk-based rules filtering. The main purpose of our framework would then be to correct the mistakes made by the former system. Consequently, reducing the workload for the AML analysts to a great extent.
\section{Experimental Evaluation}
\label{sec:experiments}

\begin{table}[]
\caption{Statistics of datasets used in the experiments. This table is an extension of Table 1 in \cite{fraudgt}.}
\label{tab:dataset-stats}
\begin{tabular}{lllll}
\hline
\textbf{Dataset} & \textbf{\# Nodes} & \textbf{\# Edges} & \textbf{Illicit Rate} & \textbf{Timespan} \\ \hline
AML Small-HI     & 0.5M              & 5M                & 0.102\%               & 10 days           \\
AML Small-LI     & 0.7M              & 7M                & 0.051\%               & 10 days           \\
AML Medium-HI    & 2M                & 32M               & 0.110\%               & 16 days           \\
AML Medium-LI    & 2M                & 31M               & 0.051\%               & 16 days           \\
AML Large-HI     & 2.1M              & 180M              & 0.124\%               & 97 days           \\
AML Large-LI     & 2M                & 176M              & 0.057\%                & 97 days           \\
ETH Phishing         & 3M                & 13.5M             & 0.270\%               & 1261 days         \\ \hline
\end{tabular}
\end{table}

In this section, we demonstrate the superior performance of {\exstraqt} with respect to accuracy and scalability, compared to the SOTA methods. The complete reproducible code for {\exstraqt} and experimental evaluation results are provided in an open access repository\footnote{https://github.com/mhaseebtariq/exstraqt}. We ran all the experiments using a machine with the following specifications, chip: Apple M3 Pro; number of cores: 12; memory: 36 GB.

\begin{table*}[tp]
\caption{Minority class F1 scores (\%) of the money laundering detection task using the AML datasets and the phishing detection task using the ETH Phishing dataset. NA stands for not available. Standard deviations are calculated over 5 runs with different random seeds. We highlight the {\color[HTML]{CB0000} \textbf{first}} and {\ul{second}} best results. GFP and MultiGNN split the train/validation/test sets with the same ratios; while FraudGT use different ratios. We report our results using both versions of splitting.}
    \label{tab:f1-table}
\begin{tabular}{l|lll|lll|l}
\hline
           & \multicolumn{3}{c|}{\textbf{AML HI}}       & \multicolumn{3}{c|}{\textbf{AML LI}}       &                    \\
\multirow{-2}{*}{\textbf{Model}} &
  \textbf{Small} &
  \textbf{Medium} &
  \textbf{Large} &
  \textbf{Small} &
  \textbf{Medium} &
  \textbf{Large} &
  \multirow{-2}{*}{\textbf{\begin{tabular}[c]{@{}l@{}}ETH \\ Phishing\end{tabular}}} \\ \hline
PNA \cite{pna}        & 56.77 ± 2.41 & 59.71 ± 1.91 & NA           & 16.45 ± 1.46 & 27.73 ± 1.65 & NA           & 51.49 ± 4.29 \\ \hline
GFP (Best) \cite{gfplib} & 64.77 ± 0.47 & 65.69 ± 0.26 & 58.03 ± 0.19 & 28.25 ± 0.80 & 31.03 ± 0.22 & 24.23 ± 0.12 & 51.00 ± 1.01       \\ \hline
FraudGT (Best) \cite{fraudgt} &
  76.41 ± 1.45 &
  75.93 ± 1.92 &
   {\ul 73.34 ± 1.64} &
  {\color[HTML]{CB0000} \textbf{47.01 ± 2.22}} &
  44.06 ± 5.27 &
  {\ul 37.43 ± 4.94} &
  NA \\ \hline
MultiGNN (Best) \cite{ppgnn} &
  68.16 ± 2.65 &
  66.48 ± 1.63 &
  61.50 ± 2.23 &
  33.07 ± 2.63 &
  36.07 ± 1.17 &
  25.35 ± 1.43 &
  {\ul 66.58 ± 1.60 } \\ \hline
ExStraQt (splits: GFP) &
  {\ul 78.90 ± 0.23 } &
  {\color[HTML]{CB0000} \textbf{80.71 ± 0.10}} &
  {\color[HTML]{CB0000} \textbf{78.05 ± 0.06}} &
  42.73 ± 0.41 &
  {\color[HTML]{CB0000} \textbf{47.67 ± 0.18}} &
  {\color[HTML]{CB0000} \textbf{38.33 ± 0.13}} &
  {\color[HTML]{CB0000} \textbf{67.03 ± 0.26}} \\
ExStraQt (splits: FraudGT) &
  {\color[HTML]{CB0000} \textbf{80.89 ± 0.32}} &
  {\ul 79.86 ± 0.15} &
  {\color[HTML]{CB0000} \textbf{78.05 ± 0.06}} &
  {\ul 45.10 ± 0.14} &
  {\ul 45.28 ± 0.06} &
  {\color[HTML]{CB0000} \textbf{38.33 ± 0.13}} &
  NA \\ \hline
\end{tabular}
\end{table*}

\subsection{Datasets}
\label{sec:datasets}

We used the following datasets in our experiments:

\subsubsection{IBM synthetic data} This dataset collection consists of six variants in two dimensions. The first dimension reflects the duration of the dataset; with \textit{small}, \textit{medium}, and \textit{large} containing 10, 16, and 97 days of data, respectively. The second dimension represents the proportion of illicit transactions; classified as low or high. The \textit{laundering} label applies to individual transactions involving suspicious funds, as well as transactions belonging to broader patterns of money laundering.

\subsubsection{Ethereum transaction network} This \textit{real-world} and \textit{timestamped} dataset \cite{ethereum1, ethereum2} represents crypto-currency exchanges between accounts on an open-source blockchain platform called Ethereum. The accounts (or nodes) involved in these transactions come with the \textit{phishing attempt} label. This label serves as a substitute for the \textit{is\_laundering} label in the IBM datasets. \cite{gfplib} and \cite{ppgnn} also use this dataset as one of their benchmarks.

Table \ref{tab:dataset-stats} shows some basic statistics related to these datasets.

\subsection{Experimental Setting}
We employ the same experimental settings as our competitors \cite{gfplib}, \cite{ fraudgt}, and \cite{ppgnn} - in terms of the evaluation metric used (F1-score); and how the datasets are split and filtered.

\subsubsection{Data split} For the sake of \textit{completeness}, we will quote the exact explanation from \cite{gfplib}: "For AML datasets, the splitting is performed such that 60\% of transactions with the smallest timestamps is selected as a training set, the next 20\% transactions with the smallest timestamps excluding the ones from the training set are selected as a validation set, and the rest are selected as the test set. For the ETH dataset, we define the timestamp of an account as the minimum timestamp among the transactions that involve this account and split the accounts of the dataset such that 65\% of the accounts with the smallest timestamp exist only in the training set, the next 15\% of the accounts exist only in the validation dataset, and the rest are in the test set. Splitting the datasets in the aforementioned way prevents data leakage in our experiments.".

The only difference from \cite{gfplib} is as follows: They report three versions for inference on the test datasets. With features generated in 1) batch (or offline) mode; 2) batch of 128 transactions; and 3) batch of 2048 transactions. Our model is not designed (although it can easily be adapted) for online training; therefore, we exclude the setups of 2) and 3) from our experiments.

Table \ref{tab:f1-table} is, in fact, an extension of the tables (number 2) in \cite{gfplib}, \cite{ppgnn}, and \cite{fraudgt}. One thing we want to stress here is that all our experiments are performed on a personal laptop, whereas for GFP \cite{gfplib}, feature generation cannot be executed for larger datasets without high amounts of available memory; for MultiGNN \cite{ppgnn} the required resources are so high that the original paper did not even report the results for IBM's large datasets; and for FraudGT \cite{fraudgt}, high compute power is required. They used an NVidia GPU for their experiments. Execution of GFP for IBM \textit{medium} dataset is not possible on a machine with 32GB of memory.

\subsection{Evaluation Results}
Table \ref{tab:f1-table} shows the exceptional results of {\exstraqt} in terms of the F1 score. Our competitors \cite{gfplib}, \cite{ppgnn}, and \cite{fraudgt} are three of the most recent works in the area. With the exception of one instance, {\exstraqt} comfortably beats all other benchmarks. \ref{tab:recall-table} shows the recall performance for {\exstraqt}, an important metric in the context of AML modeling.

{\exstraqt} outperforms GFP \cite{gfplib}, by huge margins, 100\% of the time. With uplifts in accuracy, ranging from around 20\% to a whopping 59\% on synthetic datasets. The more promising result is for the \textit{real} Ethereum dataset, where {\exstraqt} beats GFP by around 31\%. This verifies that {\exstraqt} is not only effective in detecting AML patterns in synthetic datasets, but it does an equally excellent job for a real-world application.

Compared to FraudGT \cite{fraudgt}, {\exstraqt} falls \textit{slightly} short on only one occasion. For the remaining instances, {\exstraqt} outperforms FraudGT quite comfortably. With the highest uplift in accuracy of more than 8\%. The instance where {\exstraqt} falls short; it reports a much lower variability in the output: ±0.14 compared to ±2.22 for FraudGT. This stability in output still makes {\exstraqt} much more desirable than FraudGT.

When comparing the results with MultiGNN \cite{ppgnn}, {\exstraqt} beats the benchmarks of the synthetic datasets by huge margins. The only time when MultiGNN comes close is for the (real) Ethereum dataset. Even then, {\exstraqt} reports an uplift of around 0.8\%. In addition to that, {\exstraqt} reports a much lower variability of ±0.14; compared to ±1.66 for MultiGNN.

An important thing to note here is that we did not copy all the other benchmark results from \cite{gfplib}, \cite{ppgnn}, and \cite{fraudgt}. We only picked the best performing methods from each paper. This essentially means that, by extension, {\exstraqt} beats all the unmentioned benchmarks as well.

\begin{table}[]
\caption{Mean recall scores of {\exstraqt} for the different datasets.}
\label{tab:recall-table}
\resizebox{\columnwidth}{!}{%
\begin{tabular}{|lll|lll|c|}
\hline
\multicolumn{3}{|c|}{\textbf{AML HI}} &
  \multicolumn{3}{c|}{\textbf{AML LI}} &
  \multirow{2}{*}{\textbf{\begin{tabular}[c]{@{}c@{}}Ethereum\\ Phishing\end{tabular}}} \\
\textbf{Small} &
  \textbf{Medium} &
  \textbf{Large} &
  \textbf{Small} &
  \textbf{Medium} &
  \textbf{Large} &
   \\ \hline
70.22 &
  72.45 &
  69.83 &
  30.82 &
  30.14 &
  26.13 &
  \multicolumn{1}{l|}{75.27} \\ \hline
\end{tabular}%
}
\end{table}

\subsection{Ablation Study}
\label{ref:ablation}
As part of our experiments, we performed an ablation study to analyze the positive (or negative) impact of the generated feature groups. Table \ref{tab:ablation} shows the result of that study. We can clearly see the upward trend in the F1 score, as more {\exstraqt} feature groups are added to the basic transaction features.

In the previous Section \ref{sec:anomaly}, we made the claim that the anomaly score-based features could potentially relay useful information to the model. However, in Table \ref{tab:ablation} we can see that the uplift in the F1 score is minimal. This is still not indicative of the fact that it is a useless feature group. The last column of the table proves that these features hold important information for the model, which could drastically improve the accuracy when added to the base transaction features. It is (then) a matter of finding the right combinations of feature groups based on the problem/use case at hand. Automated feature (group) selection could also be useful here. In addition, based on the predictive power of anomaly score features, they can be used as a replacement for some other (potentially) redundant features constructed on the transaction level. Consequently, the feature dimensions can be drastically reduced. The anomaly scores are estimated at the node level (with much lower volume) and then applied at the transaction level (with much higher volume). We will explore all of these areas in our future research.

\begin{table*}[tp]
\caption{Ablation study, showing the improvements in F1-score by successively adding different groups of {\exstraqt} features. The last column represents a separate variation, where only the transactions and anomaly score features are used.}
    \label{tab:ablation}
\begin{tabular}{llllll
>{\columncolor[HTML]{EFEFEF}}l }
\textbf{} &
  \textbf{transaction} &
  \textbf{+ random-walk} &
  \textbf{+ modularity} &
  \textbf{+   flows} &
  \textbf{+   anomaly score} &
  \textbf{\begin{tabular}[c]{@{}l@{}}transaction \\ + anomaly score\end{tabular}} \\
AML-Small-LI &
  0 &
  41.92 &
  42.06 &
  42.95 &
  43.02 &
  31.12 \\
AML-Small-HI &
  48.02 &
  76.93 &
  77.63 &
  78.18 &
  78.83 &
  64.36
\end{tabular}
\end{table*}

\subsection{Execution Times}
By processing huge volumes of transaction data for our experiments, on a relatively resource-constrained machine, we have (to some extent) already proved the scalability aspects of {\exstraqt}. In this section, we will report the execution times for some of the most expensive operations in {\exstraqt}.

\begin{figure}
    \centering
    \includegraphics[width=1\linewidth]{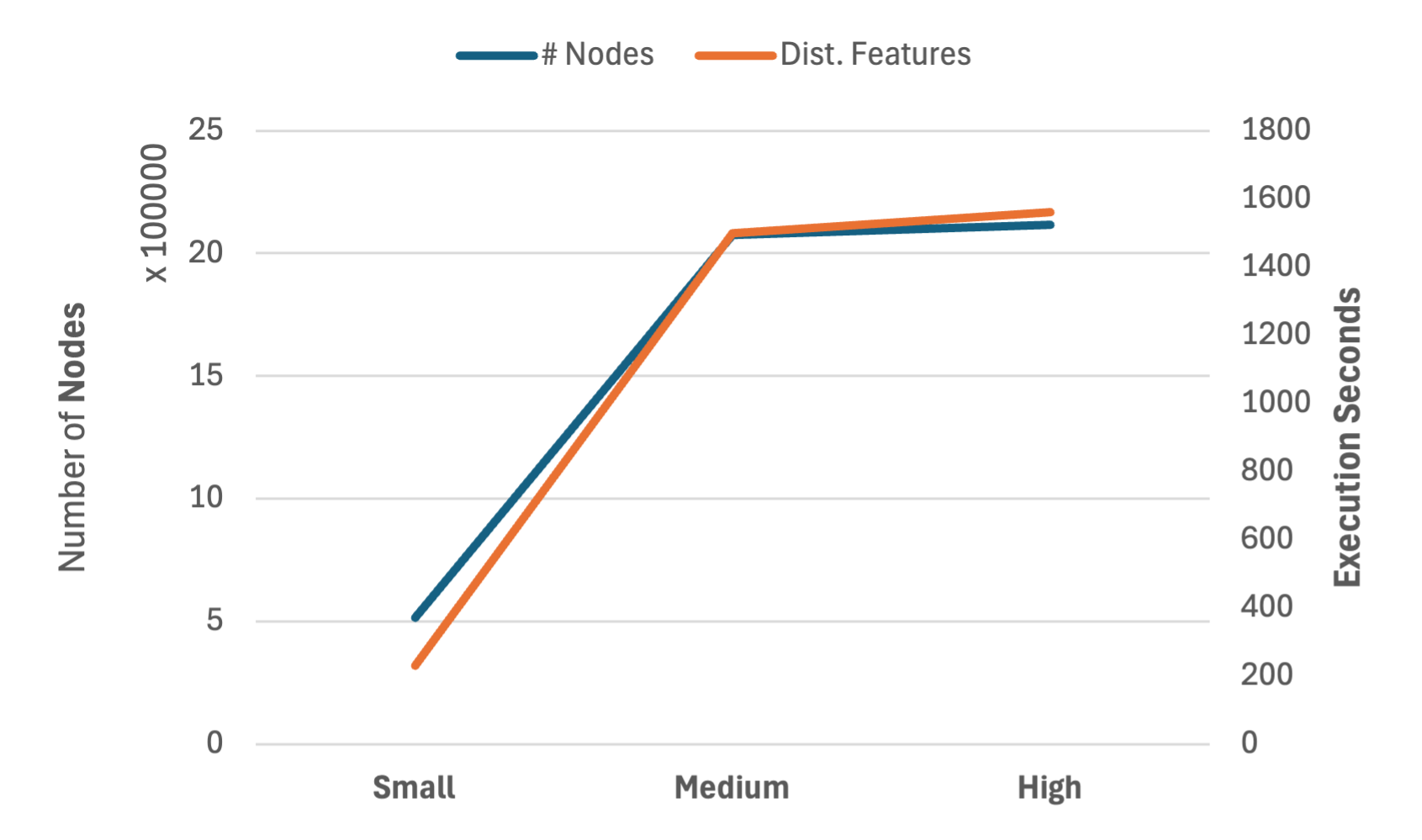}
    \caption{Execution times for the distributed graph features generation, compared to the number of nodes in the data.}
    \label{fig:exec-dist}
\end{figure}

\begin{figure}
    \centering
    \includegraphics[width=1\linewidth]{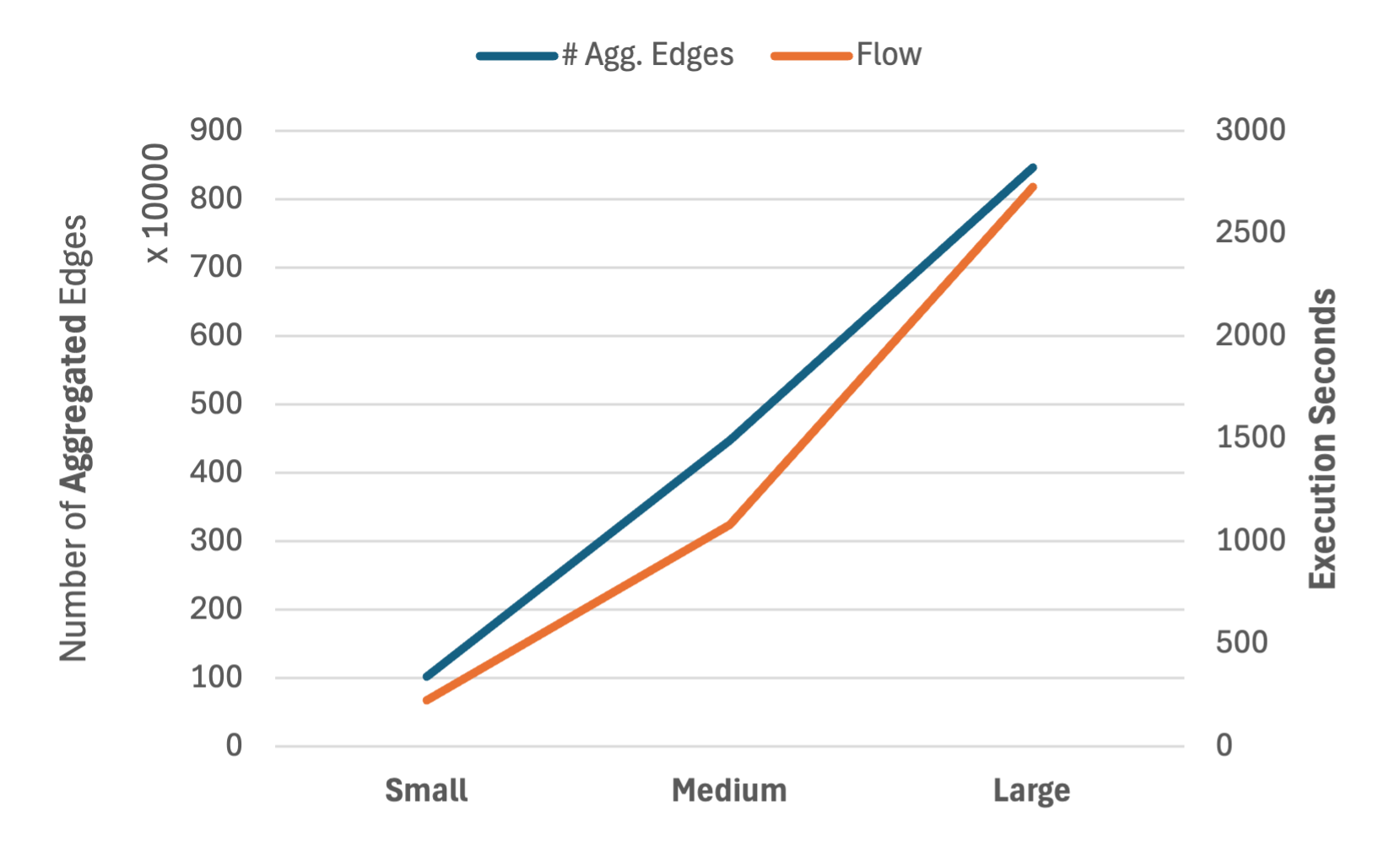}
    \caption{Execution times for the flow-based features generation, compared to the number of aggregated edges in the data.}
    \label{fig:exec-flow}
\end{figure}

\begin{figure}
    \centering
    \includegraphics[width=1\linewidth]{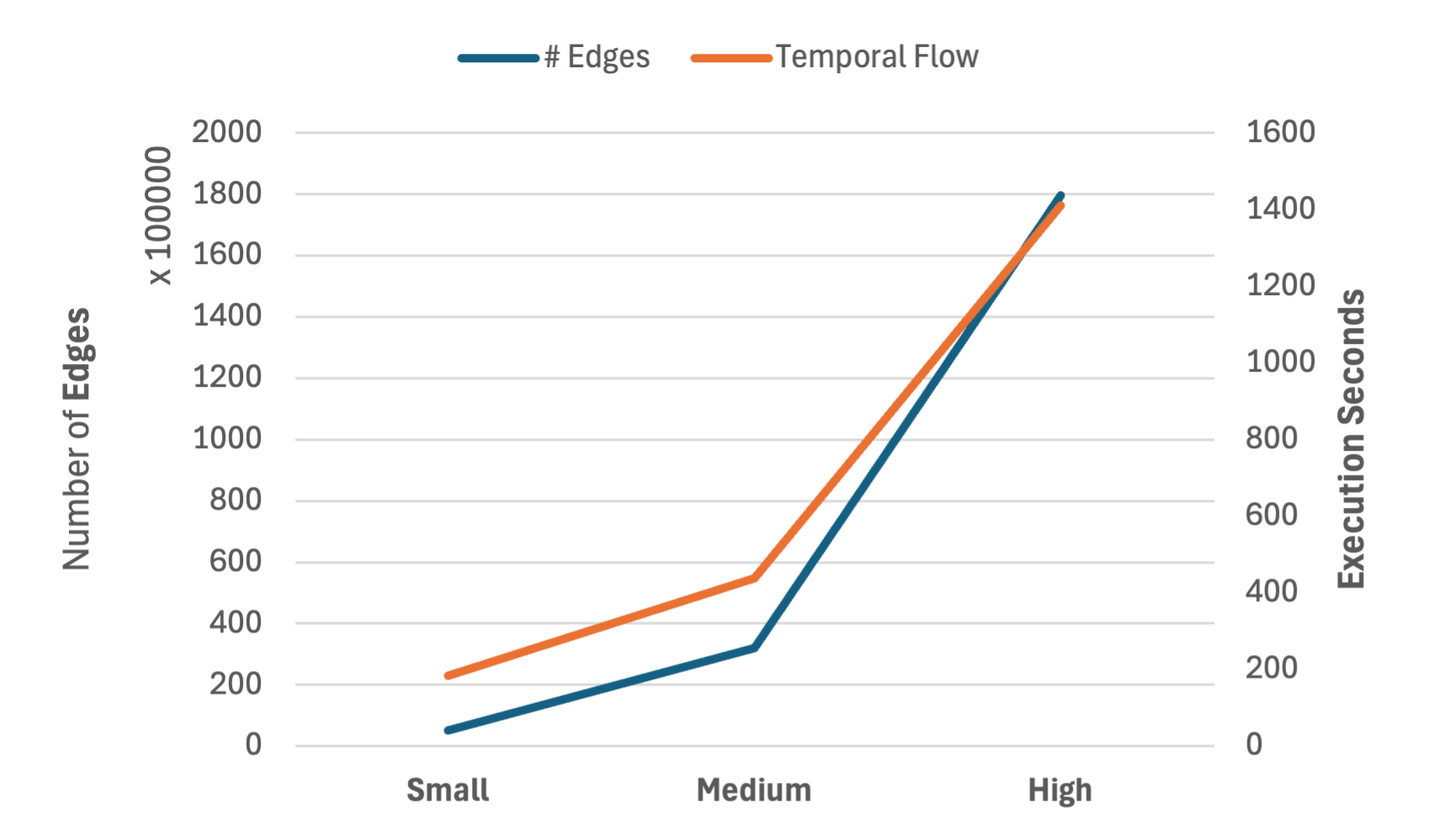}
    \caption{Execution times for the temporal flow-based features generation, compared to the number of edges (or transactions) in the data.}
    \label{fig:exec-temp}
\end{figure}

One of our main claims is that we provide a massively parallel implementation for \textit{most} of the heavy computations in {\exstraqt}. We have implemented the core components of {\exstraqt} using the distributed compute engine of PySpark \cite{spark}. The plots in Figures \ref{fig:exec-dist}, \ref{fig:exec-flow}, and \ref{fig:exec-temp} align \textit{almost} perfectly with our claims:
\begin{enumerate}
    \item Figure \ref{fig:exec-dist}: The main unit of computation for the distributed subgraph features generation logic is a \textit{node}. It then makes sense that the run time almost linearly scales with the increasing number of nodes in the datasets.
    \item Figure \ref{fig:exec-flow}: The flow-based features are generated by iteratively joining the aggregated edges. The execution time, therefore, scales with the number of aggregated edges in each dataset.
    \item Figure \ref{fig:exec-temp}: For the generation of temporal flow features, we need temporal information which is only available at the edge (or transaction) level. The implementation, therefore, scales linearly with the number of transactions (or edges).
\end{enumerate}

\begin{figure}
    \centering
    \includegraphics[width=1\linewidth]{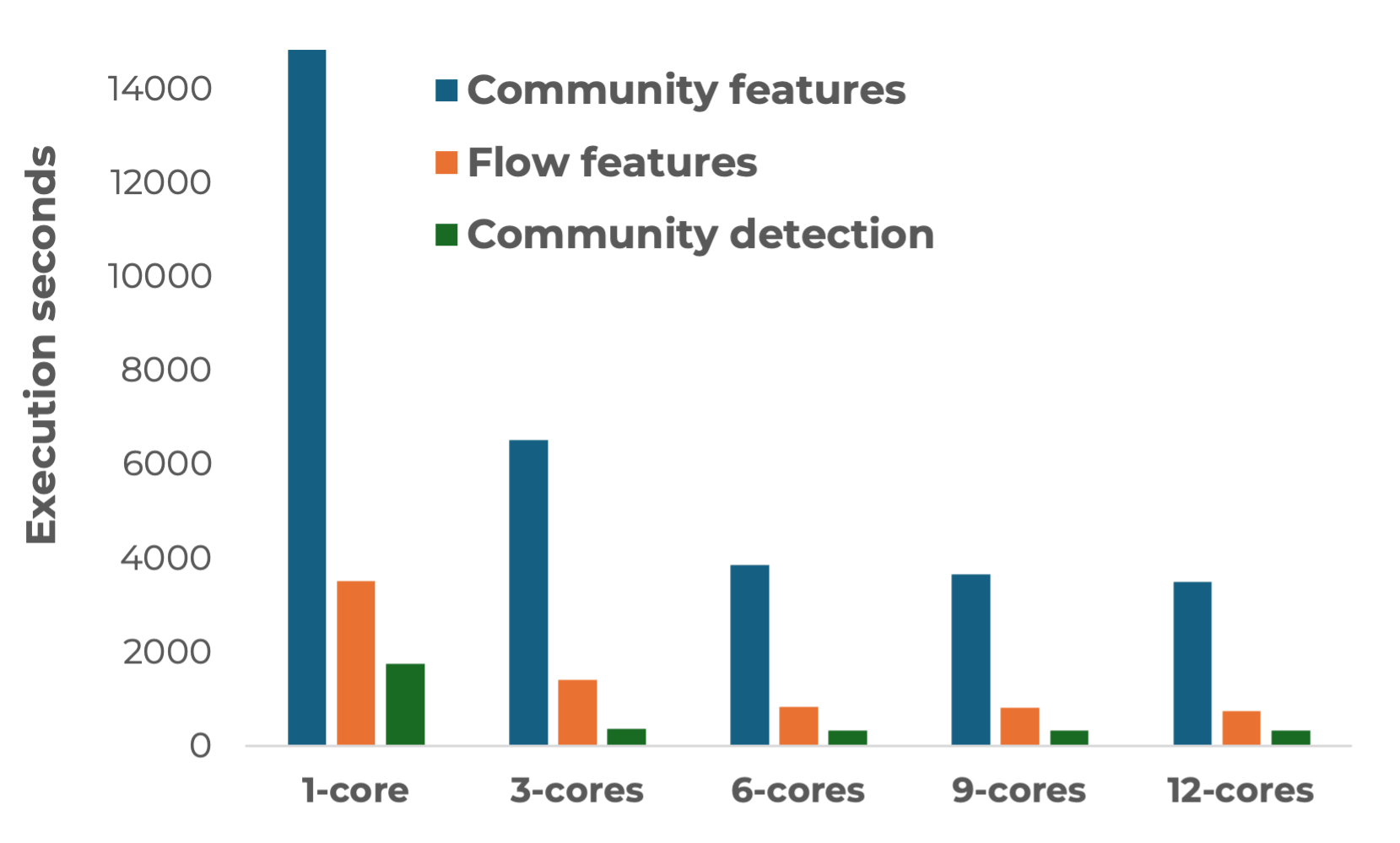}
    \caption{Execution times for different steps, with increasing level of distribution.}
    \label{fig:dist-stats}
\end{figure}

Figure \ref{fig:dist-stats} shows how the execution times for different steps can be reduced by having higher levels of parallelization. The most drastic improvement is in the execution times of community features generation. Incidentally, this is also the most time consuming task in our framework. Going from 1-core to 3-cores, we have almost a 3${\times}$ reduction in the run time. For up to 6-cores, we see substantial improvements in the run times for most tasks. After that, the (multi-) processes start fighting for the (single) machine's memory resources. The framework is written in Spark; therefore, going from multi-core to multi-node (cluster) setup is a matter of changing some configurations.
\section{Conclusion and Future Work}
\label{sec:conclusion}
We have shown that with a simple yet creative design of {\exstraqt}, a practical solution can be implemented to outperform even the most advanced Transformer and GNN-based models. The simple and scalable nature of the design of {\exstraqt} makes it a perfect candidate for an open-source AML detection framework. We encourage industry professionals to take inspiration from Section \ref{ref:prod}, for implementing {\exstraqt} in a production environment . 

With rapid advancements in GNNs, graph-based transformers, and graph auto-encoders \cite{gae}; researchers from academia and industry are investing heavily in improving and adapting these technologies for the transaction monitoring and AML detection problem spaces. However, the simpler techniques still remain underappreciated. More importantly, in highly regulated industries like financial services, there is not even an appetite for adopting the most complex models. The importance of output interpretability and model stability outweighs, to a great extent, a fractional improvement in accuracy.

We hope to continue improving and expanding {\exstraqt} in the near future. Moving forward, the following are our main areas of focus:
\begin{enumerate}
    \item Extensively explore limitations where {\exstraqt} struggle to predict well, especially compared to the SOTA models
    \item Make hyperparameter optimization and feature (group) selection, an integral and customizable part of {\exstraqt}
    \item Investigate methods to capture even better (long) \textit{temporal} relationships among suspicious actors
    \item Model ways to capture hidden ownership of \textit{common} accounts 
\end{enumerate}

\balance

\bibliographystyle{ACM-Reference-Format}
\bibliography{references}

\end{document}